\documentclass[letter, 10 pt, conference]{ieeeconf}
\IEEEoverridecommandlockouts                              
\usepackage{times}
\usepackage{epsfig}
\usepackage{graphicx}
\usepackage{amsmath}
\usepackage{amssymb}
\usepackage{multirow}

\usepackage{graphicx}
\usepackage{epsfig}
\usepackage{epic,eepic}
\usepackage{times}
\usepackage{mathptmx}
\usepackage{amsfonts}
\usepackage{amsmath}
\usepackage{cite}
\usepackage{color}

\usepackage{times}

\definecolor{red}{rgb}{1,0,0}
\definecolor{green}{rgb}{0,1,0}
\definecolor{blue}{rgb}{0,0,1}
\definecolor{violet}{rgb}{1,0,1}
\definecolor{cyan}{cmyk}{1,0,0,0}
\definecolor{magenta}{cmyk}{0,1,0,0}
\definecolor{yellow}{cmyk}{0,0,1,0}

\definecolor{white}{rgb}{1,1,1}

\usepackage{jumoline}

\newcommand{\SW}[2]{#1}

\newcommand{\CO}[2]{}

\newcommand{\CommentOut}[1]{}

 \newcommand{\editage}[1]{}

\onecolumn

\begin{document}

\newcommand{\FIG}[3]{
\begin{minipage}[b]{#1cm}
\begin{center}
\includegraphics[width=#1cm]{#2}\\
{\scriptsize #3}
\end{center}
\end{minipage}
}

\newcommand{\FIGU}[3]{
\begin{minipage}[b]{#1cm}
\begin{center}
\includegraphics[width=#1cm,angle=180]{#2}\\
{\scriptsize #3}
\end{center}
\end{minipage}
}

\newcommand{\FIGm}[3]{
\begin{minipage}[b]{#1cm}
\begin{center}
\includegraphics[width=#1cm]{#2}\\
{\scriptsize #3}
\end{center}
\end{minipage}
}

\newcommand{\FIGR}[3]{
\begin{minipage}[b]{#1cm}
\begin{center}
\includegraphics[angle=-90,clip,width=#1cm]{#2}
\\
{\scriptsize #3}
\vspace*{1mm}
\end{center}
\end{minipage}
}

\newcommand{\FIGRpng}[5]{
\begin{minipage}[b]{#1cm}
\begin{center}
\includegraphics[bb=0 0 #4 #5, angle=-90,clip,width=#1cm]{#2}\vspace*{1mm}
\\
{\scriptsize #3}
\vspace*{1mm}
\end{center}
\end{minipage}
}

\newcommand{\FIGpng}[5]{
\begin{minipage}[b]{#1cm}
\begin{center}
\includegraphics[bb=0 0 #4 #5, clip, width=#1cm]{#2}\vspace*{-1mm}\\
{\scriptsize #3}
\vspace*{1mm}
\end{center}
\end{minipage}
}

\newcommand{\FIGtpng}[5]{
\begin{minipage}[t]{#1cm}
\begin{center}
\includegraphics[bb=0 0 #4 #5, clip,width=#1cm]{#2}\vspace*{1mm}
\\
{\scriptsize #3}
\vspace*{1mm}
\end{center}
\end{minipage}
}

\newcommand{\FIGRt}[3]{
\begin{minipage}[t]{#1cm}
\begin{center}
\includegraphics[angle=-90,clip,width=#1cm]{#2}\vspace*{1mm}
\\
{\scriptsize #3}
\vspace*{1mm}
\end{center}
\end{minipage}
}

\newcommand{\FIGRm}[3]{
\begin{minipage}[b]{#1cm}
\begin{center}
\includegraphics[angle=-90,clip,width=#1cm]{#2}\vspace*{0mm}
\\
{\scriptsize #3}
\vspace*{1mm}
\end{center}
\end{minipage}
}

\newcommand{\FIGC}[5]{
\begin{minipage}[b]{#1cm}
\begin{center}
\includegraphics[width=#2cm,height=#3cm]{#4}~$\Longrightarrow$\vspace*{0mm}
\\
{\scriptsize #5}
\vspace*{8mm}
\end{center}
\end{minipage}
}

\newcommand{\FIGf}[3]{
\begin{minipage}[b]{#1cm}
\begin{center}
\fbox{\includegraphics[width=#1cm]{#2}}\vspace*{0.5mm}\\
{\scriptsize #3}
\end{center}
\end{minipage}
}

\newcommand{\figAa}[1]{
\FIGpng{3}{all-picture3/#1-1.png}{}{256}{256}\hspace*{-1mm}%
\FIGpng{4.3}{all-picture3/#1-2.png}{}{640}{480}\hspace*{-11mm}%
\FIGpng{4.3}{all-picture3/#1-3.png}{}{640}{480}\hspace*{-11mm}%
\FIGpng{4.3}{all-picture3/#1-4.png}{}{640}{480}\hspace*{-11mm}%
\FIGpng{3}{all-picture3/#1-5.png}{}{256}{256}\\
}

\newcommand{\figA}{
\begin{figure*}[t]
\begin{center}
\figAa{1}
\figAa{2}
\figAa{3}
\figAa{4}
\figAa{5}
\caption{
Examples of image change detection. Each row, from left to right, shows the query scene, the query image, the aligned query-reference image pair in invariant coordinate system, the reference image, and the reference scene.
}\label{fig:a}
\end{center}
\end{figure*}
}

\newcommand{\figB}{
\begin{figure*}[t]
\begin{center}
\FIG{4}{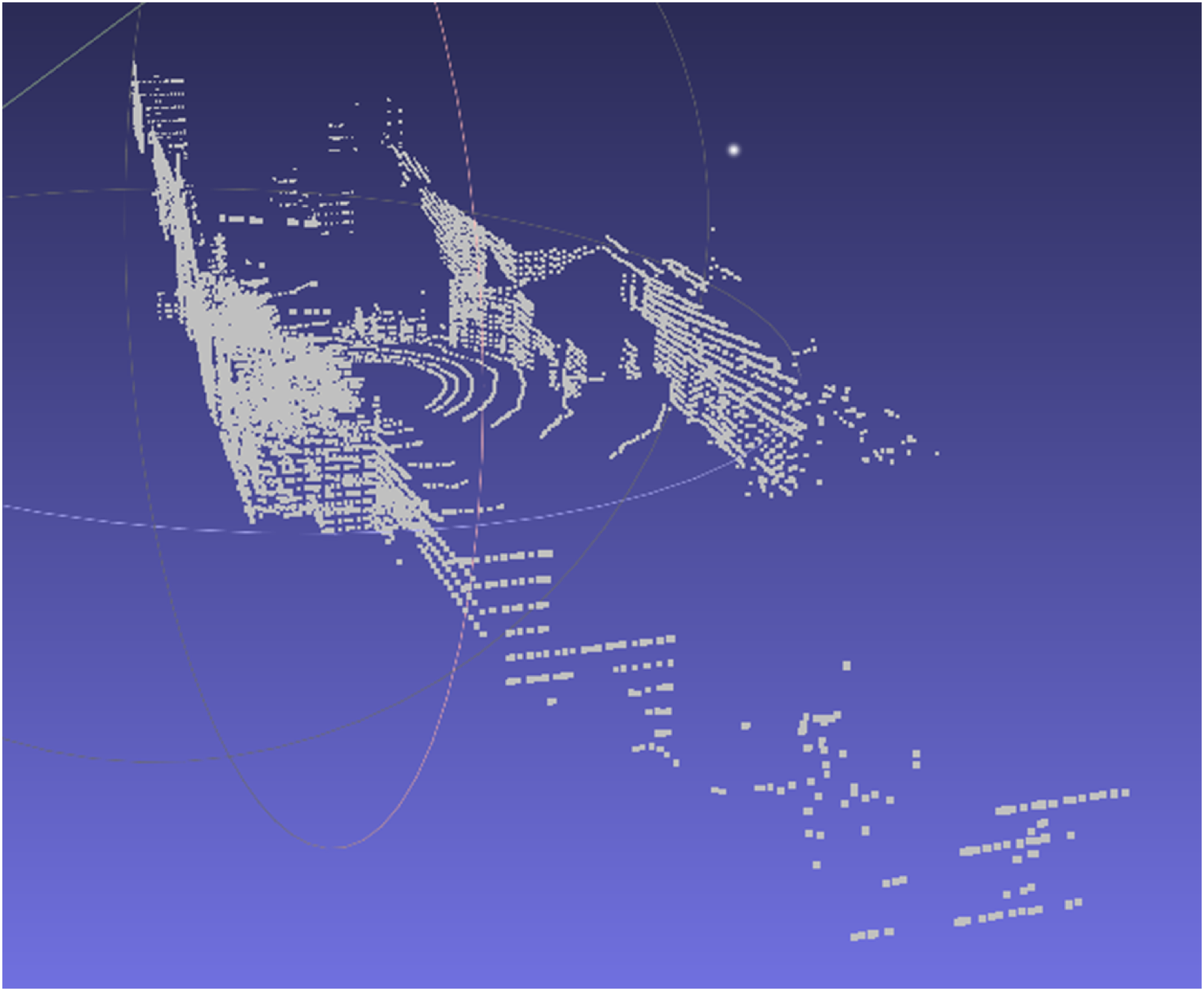}{a}
\FIG{3.5}{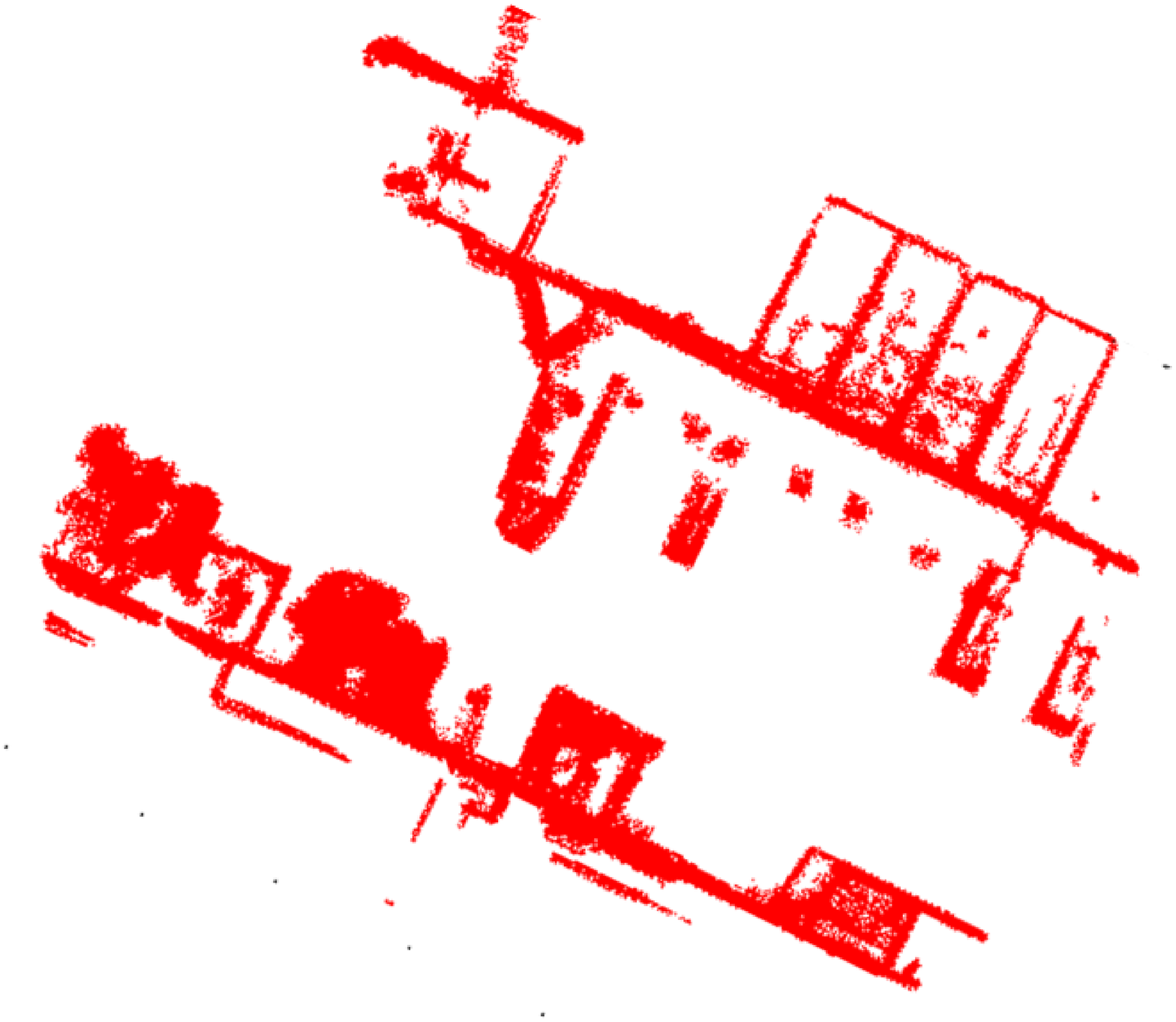}{b}
\FIG{3.5}{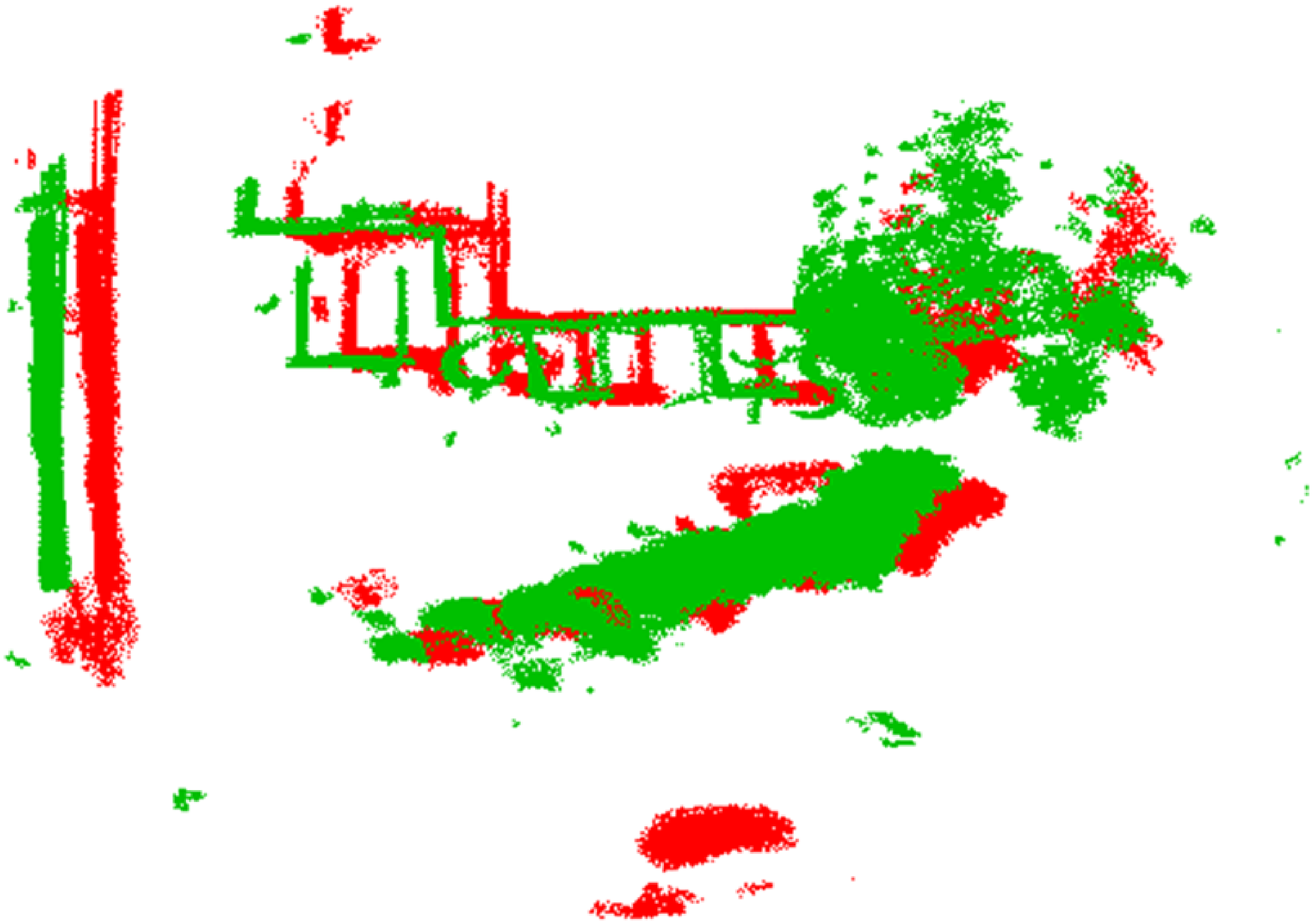}{c}
\FIG{2.5}{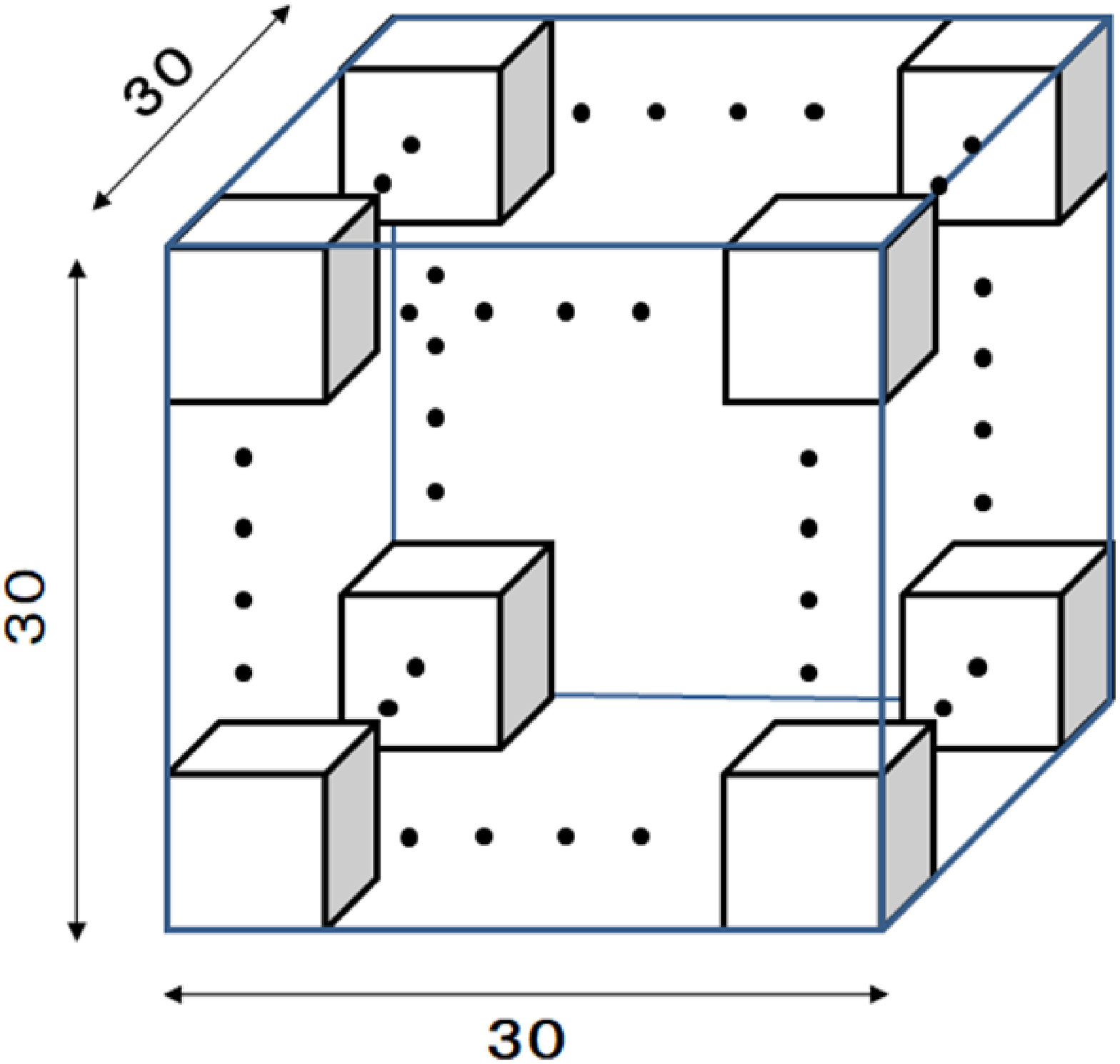}{d}
\FIG{2.5}{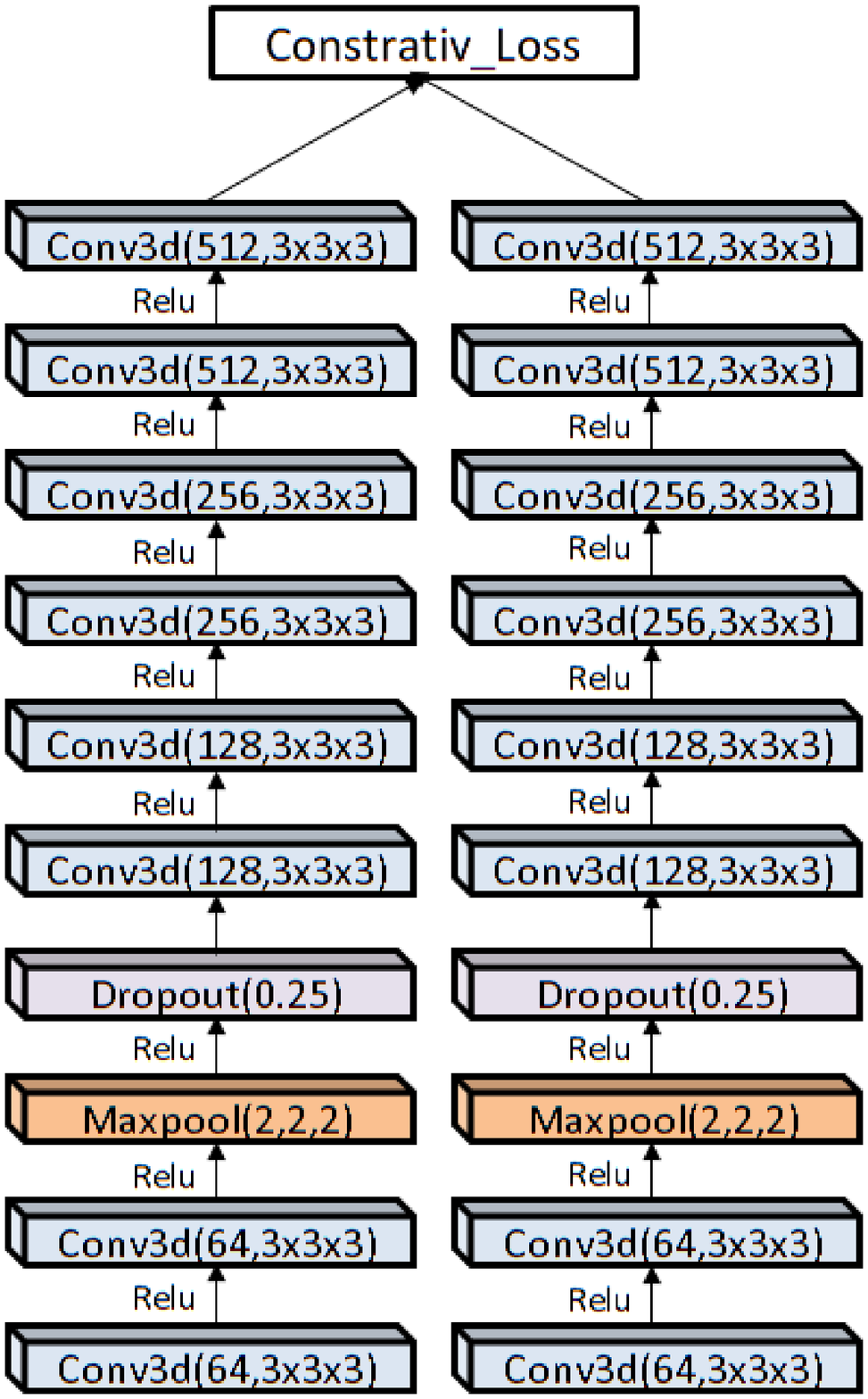}{e}
\caption{Proposed framework. (a) A short sequence of 3D scans is acquired by the on-board imagery. (b) A 3D image (local map) is built by aligning scans in the input sequence. (c) Every query/reference image is aligned with the invariant coordinate system to allow direct comparison of the 3D keypoints. (d) A constant-dimensional truncated distance function (TDF) vector is obtained from each of the local interest regions in a query/reference image. (e) A Siamese network is trained from weakly labeled TDF vector pairs to learn discriminative 3D neural codes.
}\label{fig:b}
\end{center}
\end{figure*}
}

\newcommand{\figD}{
\begin{figure}[t]
\FIGpng{8}{cd18_fig4.png}{}{5168}{8380}
\SW{
\caption{
Examples of LoC images.
From left to right,
each panel shows
an input image,
the LoC image by the proposed ``rank fusion" method,
the LoC image by the ``score sum",
and the LoC image by the ``score max".
}\label{fig:examples}
}{
}
\end{figure}
}

\newcommand{\figE}{
\begin{figure}[t]
\FIGpng{4}{fig1/1327251047113497.png}{}{1232}{1616}%
\FIGpng{4}{fig1/1333220460511895.png}{}{1232}{1616}\vspace*{-5mm}\\
\FIGpng{4}{fig1/pc0-15-c.png}{}{500}{330}%
\FIGpng{4}{fig1/pc0-17-c.png}{}{550}{330}\vspace*{-5mm}\\
\FIGpng{4}{fig1/pc2-15-c.png}{}{550}{330}%
\FIGpng{4}{fig1/pc2-17-c.png}{}{550}{330}
\caption{Cross-season change retrieval (CS-CR).
Left: query scene. Right: reference scene.
Each column shows,
from top to bottom,
visual images (for visualization only and not used in experiments),
3D point cloud images aligned with the invariant coordinate system (ICS),
and
discriminatively learned deep 3D neural codes 
(different colors in the bottom panel
correspond to different visual words).
}\label{fig:e}
\end{figure}
}

\newcommand{\vs}{\hspace*{-1mm}}

\newcommand{\tabA}{
\begin{table}[t]
\begin{center}
\caption{Performance results.}\label{tab:a}
\begin{tabular}{|l|r|r|r|}
\hline
\multirow{2}{2cm}{} & VL & \multicolumn{2}{c|}{ICD} \\ \cline{2-4}
 & & w/o ICS & w/ ICS \\ \hline
FPFH & 15.1 & - & - \\
Siamese & 10.8 & - & - \\
FPFH-BoW & 39.3 & 32.9 & 52.1 \\
Siamese-BoW & 13.1 & 71.0 & 83.0 \\ \hline
\end{tabular}
\end{center}
\end{table}
}

\renewcommand{\CO}[2]{#2}



\title{\LARGE \bf
Scalable Change Retrieval 
Using
Deep 3D Neural Codes
}
\author{Kojima Yusuke ~~~~ Tanaka Kanji ~~~ Yang Naiming ~~~ Hirota Yuji
\thanks{Our work has been supported in part by 
JSPS KAKENHI 
Grant-in-Aid 
for Scientific Research (C) 26330297, and (C) 17K00361.}
\thanks{The authors are with Graduate School of Engineering, University of Fukui, Japan. 
{\tt\small tnkknj@u-fukui.ac.jp}}
}

\maketitle

\begin{abstract} 
We present a novel scalable framework for image change detection (ICD) from an on-board 3D imagery system. We argue that existing ICD systems are constrained by the time required to align a given query image with individual reference image coordinates. We utilize an invariant coordinate system (ICS) to replace the time-consuming image alignment with an offline pre-processing procedure. Our key contribution is an extension of the traditional image comparison-based ICD tasks to setups of the image retrieval (IR) task. We replace each component of the 3D ICD system, i.e., (1) image modeling, (2) image alignment, and (3) image differencing, with significantly efficient variants from the bag-of-words (BoW) IR paradigm. Further, we train a deep 3D feature extractor in an unsupervised manner using an unsupervised Siamese network and automatically collected training data. We conducted experiments on a challenging cross-season ICD task using a publicly available dataset and thereby validate the efficacy of the proposed approach.
\end{abstract}

\section{Introduction}

Recent progress in simultaneous localization and mapping (SLAM) has led to the development of various practical SLAM systems (e.g., LSD-SLAM \cite{lsdslam}) that can map large environments via egocentric on-board 3D vision (e.g., 3D LIDAR imagery). However, SLAM in non-stationary environments remains a significant challenge \cite{SeqSlamOrg,churchill2013experience,paton2016bridging,iv18mapmaintenance}. A major source of difficulty is the large number of possible changes between live images and a map (e.g., of car parking and building construction), which grows combinatorially relative to the map size. To reduce computational and perceptual complexity, it is helpful if an SLAM system has the ability of detecting changed objects.

The detection of changed objects in a query live image relative to a pre-built background model (i.e., a map) is a fundamental problem in computer vision called image change detection (ICD) \cite{icdsurvey}, which has been studied in many different contexts, including remote sensing \cite{remotesensing} and surveillance \cite{surveillance}. In these classical contexts, the problem is typically formulated as a two-stage image comparison process. The first stage, alignment, is aimed to align the query image with the reference image coordinate system. The second stage, differencing, is aimed to compare the query and reference images with respect to the reference coordinate system. However, this two-stage process requires a large number of alignment and differencing operations for every possible reference image, the time cost of which is prohibitive for time-critical vehicular applications.

We were particularly interested in the use of the viewpoint invariant coordinate system (ICS) \cite{itsc18takahashi}, which solves the above problems, and this motivated our study. Our approach integrates the alignment and differencing steps in a unique manner. Our key idea is to align an input query image with a pre-defined ICS, rather than to align it with individual reference images. This allows the image alignment to be a part of the offline pre-processing, which leads to a significant reduction in online processing \cite{itsc18lmd}. 

The main contribution of this paper is an extension of the traditional image comparison task to setups of the image retrieval task (Fig. \ref{fig:b}). More formally, we replace each component of the 3D ICD system, i.e., (1) image modeling, (2) image alignment, and (3) image differencing, with significantly efficient variants from the bag-of-words (BoW) image retrieval paradigm \cite{harris1954distributional}. Our BoW-based approach was primarily motivated by two independent fields: viewpoint-localization (VL) \cite{fabmap2} and ICD \cite{bowicd}, in which codeword (i.e., BoW) models are used as compact appearance models based on vector quantization. Our key contribution is that the advantages of the two independent BoW techniques are combined into a unified change retrieval (CR) framework. We introduce the ICS mentioned above to allow the direct comparison of appearance features in the query and reference images without the online pre-alignment step. Further, we train a deep 3D feature extractor in an unsupervised manner using an unsupervised Siamese network and a weakly labeled training set automatically collected from the target environments. The result is an extremely efficient deep CR framework that requires only a single look-up of the inverted file and a few L2 norm comparisons. The results of experiments on a challenging cross-season CR (CS-CR) task using the North Campus Long-Term (NCLT) dataset \cite{nclt}  validate the feasibility of the proposed method.

\figB

\section{Approach}

The objective of CR is to detect a significant change between query and reference images that are observed at different times. A query image is modeled as a local point-cloud (PC) map, which is built from a sequence of sensor data acquired by a vehicle's on-board 3D scanner during short-range navigation. A reference image is modeled by a local PC image in the same format. Each short sequence corresponds to the vehicle's travel distance of 5 m. A local map is constructed by aligning this short sequence of 3D scans. The alignment is based on dead-reckoning, inertial measuring units, and an iterative closest point (ICP) algorithm.

We followed studies in the literature \cite{icdsurvey} to formulate the CR task as a task consisting of localizing change objects with respect to the environment. Unlike many other applications, vehicular applications cannot always assume the availability of precise viewpoint information. For example, GPS is frequently occluded by trees, terrain, or buildings. Therefore, we must consider the task of joint viewpoint-change localization under global viewpoint uncertainty \cite{itsc18takahashi}.

The performance of a retrieval task is frequently evaluated by a ranking-based metric. In this study, we employed two different such metrics, averaged normalized rank (ANR) \cite{icra15ando} and top-$X$ accuracy \cite{faster-r-cnn}, for the VL and ICD tasks, respectively. ANR utilizes a ranked list of viewpoints in descending order of relevance score. It is defined as the average of the ranks of the ground-truth viewpoint normalized by the database size. Top-$X$ accuracy utilizes a collection of top-$X$ datapoints with the highest likelihood of change (LoC) and is defined as the ratio of query images where ICD is successful. The success of ICD is defined by whether the ground-truth bounding boxes (of changed objects) overlap with the top-$X$ ranked datapoints.

The proposed CR framework consists of offline and online processes. 
Offline, a discriminative feature extractor is trained in an unsupervised manner (Section \ref{sec:extractor}) from a weakly-labeled training set. 
Online, first each of the images is converted to a collection of local features (Section \ref{sec:extractor}), and then, their appearance descriptors are translated to the compact BoW representation (Section \ref{sec:coding}), while their 3D keypoints are transformed to ICS (Section \ref{sec:ics}).
Then, VL (Section \ref{sec:localization}) and ICD (Section \ref{sec:cd}) are performed based on the BoW and ICS formulations.
These processes are detailed in the following subsections.

\subsection{Three-dimensional Deep Neural Codes}\label{sec:extractor}

The feature extractor is aimed to extract a collection of local features from a given query/reference image. This task consists of three stages. First, a collection of $N=500$ is sampled from the PC. Then, the local interest region of each keypoint is 
translated to a $30^3$ dimension vector 
by the truncated distance function (TDF) \cite{zeng20163dmatch}. 
Finally, the TDF vector is translated 
to a 512-dim discriminative feature vector
by a convolutional neural network (CNN). The TDF and CNN are detailed in the following. 

The TDF is a technique for translating variable size PC data to a constant dimension vector representation. 
This constant dimension vector representation is useful for inputting data to a CNN, as most state-of-the-art CNNs are based on the assumption of a constant dimension input. In our approach, a size 30$\times$30$\times$30 voxel grid is imposed within a given 9$\times$9$\times$9 [$m^3$] interest region, and the local PC in each voxel is approximated by its TDF value, which yields a $30^3=27,000$ dimensional vector.

The CNN is pre-trained in an unsupervised manner by using a Siamese network.
A Siamese network is a Y-shaped neural network that joins two network branches in the final layers to produce a single output. 
The two branches have the same layer structure. 
The network is trained in a weakly-supervised scheme for learning a similarity measure from training samples of similar/dissimilar TDF vector pairs. Each branch is a CNN with eight 3D convolutional layers. 
These layers have different numbers of kernels of 64-64-128-128-256-256-512-512 with a kernel size of 3$\times$3. A max-pooling layer with a pool size of 2$\times$2$\times$2 and a dropout layer of ratio 0.25 are inserted after the second convolutional layers. 
Inspired by \cite{zeng20163dmatch}, 
we consider the trained CNN a discriminative feature extractor 
and use it as a feature extractor for converting a given TDF vector to a 512 dimensional feature vector. 
We extend 
the framework
of \cite{zeng20163dmatch}
from
indoor RGB-D data 
to outdoor 3D scanner data.

The training data for the feature extractor can be automatically collected 
by the vehicle itself
in the target domain. 
A collection of similar/dissimilar TDF vector pairs is required as positive/negative training data for the Siamese network. 
The positive samples are defined as TDF vector pairs 
(from the training sequence)
in two different viewpoints 
originating from the same real-world object. 
The negative samples are the other majority of TDF vector pairs.
Therefore,
negative samples can be easily sampled from the TDF vector pairs.
However, 
the automatic collection of positive pairs is not a trivial task. 
In this study, we employed a two stage approach. 
First, relative localization between two successive local PC images at time stamp $t$ and $t+3$ [sec] 
within the training sequence
is performed by using the available dead-reckoning measurements. 
Second, fine-grained registration is performed 
using the ICP as described in \cite{icp}. 
Then, the two aligned PCs are both cropped 
by 
the same bounding box of
size
0.3$\times$0.3$\times$0.3 $m^3$ 
at the interest region
to obtain paired TDF vectors. 
It is noteworthy 
that
the above procedure is fully automatic and does not require human intervention.

\subsection{Bag of Words Representation}\label{sec:coding}

Local features in query/reference images are encoded to the BoW representation. Formally, each local feature is vector quantized 
and encoded
to visual word $w \in [1,W]$ ($W=10,000$). This is equivalent to the task of finding the nearest neighbor (NN) exemplar in a pre-defined collection of exemplar features called a vocabulary, and the ID of the NN exemplar is defined as the visual word. Then, the visual word can be used as a compact index for efficient image retrieval. For the vocabulary, a classifier with $W$ clusters is pretrained offline using the unsupervised k-means clustering algorithm presented in \cite{macqueen1967some}. The training image 
for the classifier
is a size 
1.2M collection of local features
(from 2,409 images), which is independent of the training/testing data used for experiments.

\subsection{Invariant Coordinate System (ICS)}\label{sec:ics}

The keypoints of each local feature 
in each query/reference image
are transformed to the ICS \cite{itsc18takahashi}.
The ICS should be 
designed to be invariant to the vehicle viewpoints, 
to allow direct comparison of the keypoints of local features
in different 3D images
under viewpoint uncertainty. 
It is noteworthy that 
online alignment is required only for the query image. The alignment for reference images 
can be accomplished offline, which leads to 
a significant reduction in online computation.

A key design issue is 
the determination of the origin and axes of the ICS \cite{itsc18lmd}. 
In this  study,
we assumed that the vertical axis
of the image coordinate is known
and orthogonal to the horizontal ground plane.
For determining the 
two horizontal 
coordinate axes, we adopted the entropy minimization criteria described in \cite{olufs2011robust}. For determining the origin, we adopted a strategy called ``center-of-gravity (CoG)", which determines the origin as the centre of gravity of the given PC. 
The CoG strategy was proposed in our previous paper \cite{sii2014}, further verified in \cite{IPIN17}, and successfully applied in tasks on VL \cite{DBLP:conf/itsc/TakahashiTF17} and ICD \cite{itsc18lmd}. Moreover, the current paper also describes an ablation study to verify the effectiveness of ICS in our novel CR task.

\subsection{Viewpoint Localization}\label{sec:localization}\label{sec:vl}

3D BoW features are fed to a standard VL algorithm,
in which 
the NN reference image $j$ is determined by
minimizing the naive Bayes nearest neighbor (NBNN)
distance metric:
\begin{equation}
j = arg \min_j \sum_{q\in Q} \min_{r\in R_j} | q - E(r)|.
\end{equation}
$|\cdot|$ is L2 norm.
$Q$ and $R_j$ are the collections of features from the query and the $j$-th reference image,
respectively. 
$q$ is a raw feature vector in the query image.
$E(r)$ is an
exemplar that corresponds to the visual word of interest
and 
approximates
the reference feature vector $r$
(Section \ref{sec:coding}) 
that 
corresponds to the 
visual word
of interest.

\subsection{Image Change Detection}\label{sec:cd}

The relevant reference images hypothesized by the VL 
are 
compared to 
a query image 
in terms of appearance and spatial cues. Appearance cues are the BoW representations that 
are 
extracted by 
a discrimatively trained CNN (Section \ref{sec:extractor}) and 
further encoded to the 
BoW-based representation (Section \ref{sec:coding}). 
Spatial cues are the 3D keypoints that are based on ICS (Section \ref{sec:ics}). 
Based on these appearance and spatial cues,
the LoC
of a datapoint in a given query image
is defined as
the L2 distance to the NN
reference image
in the feature space.
Following \cite{spm},
its search region 
for the NN reference feature 
is defined as a horizontal bounding box
in ICS.
The horizontal plane is partitioned 
at 
$x$=0, $\pm$$\bar{x}$ on $x$-axis
and
at
$y$=0, $\pm$$\bar{y}$ on $y$-axis, 
into 
4$\times$4 grid cells
and those cell
which the query keypoint belongs to 
is considered the local search region.
$\bar{x}$ and $\bar{y}$ are means of the absolute values of the $x$ and $y$ of all the datapoints in the training images.

\tabA

\section{Experiments and Discussions}

We evaluated the suitability of the methods presented above for CS-CR using the  NCLT dataset \cite{nclt}. 

This dataset is a long-term autonomy dataset for robotics research collected at the University of Michigan's North Campus. The dataset consists of omnidirectional imagery, 3D LIDAR, planar LIDAR, GPS, and odometry data. 
During vehicle travel in outdoor environments, various types of appearance changes were encountered with respect to the data from different seasons. These changes originate from the movement of people, parked cars, building construction, and other nuisance changes originating from viewpoint-dependent changes of object appearances and occlusions, weather changes, falling leaves, and snow. These appearance changes make our CS-CR task challenging. 

We utilized the PC data from the LIDAR as input to our CS-CR tasks. We used the available GPS information only as the ground-truth data.
We used datasets of four different seasons, where the images were captured on ``2012/3/31 (SP)," ``2012/8/04 (SU)," ``2012/11/17 (AU)," and ``2012/1/22 (WI)," as individual training/testing sets. As aforementioned, we did not assume the availability of ground-truth viewpoint information, and considered multiple candidates of relevant reference images. The numbers of these reference image candidates were 130, 116, 121, and 133 for 
SP, SU, AU, and WI.

We created an image set consisting of 100 query images. More specifically, we considered all the 12 different combinations of query-reference dataset seasons. Each query image belonged to a different season's dataset from those of the reference images. In each selected query image, the ground-truth GPS locations of the two images were sufficiently close and newly appearing change objects (e.g., parking cars) were present with respect to the relevant reference image. We annotated the changed objects in the query image by 2D bounding boxes on the horizontal plane.
We compared the proposed ICD algorithm using deep neural codes with an alternative algorithm based on the fast pont feature histogram (FPFH) presented in \cite{rusu2009fast}.

Table \ref{tab:a} shows the ICD performance in terms of top-$X$ accuracy. 
The proposed method frequently successfully aligned the invariant coordinate system of the query and the reference images. This allowed the spatial information of local features with respect to the coordinate systems to be utilized as an additional discriminative cue for VL. The proposed method clearly outperforms the baseline FPFH method.

\bibliographystyle{IEEEtran}
\bibliography{cite}

\begin{thebibliography}{10}
\providecommand{\url}[1]{#1}
\csname url@rmstyle\endcsname
\providecommand{\newblock}{\relax}
\providecommand{\bibinfo}[2]{#2}
\providecommand\BIBentrySTDinterwordspacing{\spaceskip=0pt\relax}
\providecommand\BIBentryALTinterwordstretchfactor{4}
\providecommand\BIBentryALTinterwordspacing{\spaceskip=\fontdimen2\font plus
\BIBentryALTinterwordstretchfactor\fontdimen3\font minus
  \fontdimen4\font\relax}
\providecommand\BIBforeignlanguage[2]{{%
\expandafter\ifx\csname l@#1\endcsname\relax
\typeout{** WARNING: IEEEtran.bst: No hyphenation pattern has been}%
\typeout{** loaded for the language `#1'. Using the pattern for}%
\typeout{** the default language instead.}%
\else
\language=\csname l@#1\endcsname
\fi
#2}}

\bibitem{lsdslam}
J.~Engel, T.~Sch{\"o}ps, and D.~Cremers, ``Lsd-slam: Large-scale direct
  monocular slam,'' in \emph{European Conference on Computer Vision}.\hskip 1em
  plus 0.5em minus 0.4em\relax Springer, 2014, pp. 834--849.

\bibitem{SeqSlamOrg}
M.~J. Milford and G.~F. Wyeth, ``Seqslam: Visual route-based navigation for
  sunny summer days and stormy winter nights,'' in \emph{Robotics and
  Automation (ICRA), 2012 IEEE International Conference on}.\hskip 1em plus
  0.5em minus 0.4em\relax IEEE, 2012, pp. 1643--1649.

\bibitem{churchill2013experience}
W.~Churchill and P.~Newman, ``Experience-based navigation for long-term
  localisation,'' \emph{The International Journal of Robotics Research},
  vol.~32, no.~14, pp. 1645--1661, 2013.

\bibitem{paton2016bridging}
M.~Paton, K.~MacTavish, M.~Warren, and T.~D. Barfoot, ``Bridging the appearance
  gap: Multi-experience localization for long-term visual teach and repeat,''
  in \emph{Intelligent Robots and Systems (IROS), 2016 IEEE/RSJ International
  Conference on}.\hskip 1em plus 0.5em minus 0.4em\relax IEEE, 2016, pp.
  1918--1925.

\bibitem{iv18mapmaintenance}
B.~Mathias, D.~Marcin, G.~Igor, C.~Cesar, S.~Roland, and N.~Juan, ``Map
  management for efficient long-term visual localization in outdoor
  environments,'' in \emph{Intelligent Vehicle Symposium, IEEE}, 2018.

\bibitem{icdsurvey}
R.~J. Radke, S.~Andra, O.~Al-Kofahi, and B.~Roysam, ``Image change detection
  algorithms: a systematic survey,'' \emph{IEEE transactions on image
  processing}, vol.~14, no.~3, pp. 294--307, 2005.

\bibitem{remotesensing}
L.~Gueguen and R.~Hamid, ``Large-scale damage detection using satellite
  imagery,'' in \emph{Proceedings of the IEEE Conference on Computer Vision and
  Pattern Recognition}, 2015, pp. 1321--1328.

\bibitem{surveillance}
W.~Sultani, C.~Chen, and M.~Shah, ``Real-world anomaly detection in
  surveillance videos,'' \emph{Center for Research in Computer Vision (CRCV),
  University of Central Florida (UCF)}, 2018.

\bibitem{itsc18takahashi}
\BIBentryALTinterwordspacing
Y.~Takahashi, K.~Tanaka, and N.~Yang, ``Scalable change detection from 3d point
  cloud maps: Invariant map coordinate for joint viewpoint-change
  localization,'' in \emph{21st International Conference on Intelligent
  Transportation Systems, {ITSC} 2018, Maui, HI, USA, November 4-7, 2018},
  2018, pp. 1115--1121. [Online]. Available:
  \url{https://doi.org/10.1109/ITSC.2018.8569294}
\BIBentrySTDinterwordspacing

\bibitem{itsc18lmd}
------, ``Scalable change detection from 3d point cloud maps: Invariant map
  coordinate for joint viewpoint-change localization,'' in \emph{Intelligent
  Transportation Systems (ITSC), 2018 IEEE 21th International Conference on},
  2018.

\bibitem{harris1954distributional}
Z.~S. Harris, ``Distributional structure,'' \emph{Word}, vol.~10, no. 2-3, pp.
  146--162, 1954.

\bibitem{fabmap2}
M.~Cummins and P.~M. Newman, ``Appearance-only {SLAM} at large scale with
  {FAB-MAP} 2.0,'' \emph{I. J. Robotics Res.}, vol.~30, no.~9, pp. 1100--1123,
  2011.

\bibitem{bowicd}
K.~Kim, T.~H. Chalidabhongse, D.~Harwood, and L.~Davis, ``Real-time
  foreground--background segmentation using codebook model,'' \emph{Real-time
  imaging}, vol.~11, no.~3, pp. 172--185, 2005.

\bibitem{nclt}
N.~Carlevaris-Bianco, A.~K. Ushani, and R.~M. Eustice, ``University of michigan
  north campus long-term vision and lidar dataset,'' \emph{The International
  Journal of Robotics Research}, pp. 1023--1035, 2015.

\bibitem{icra15ando}
\BIBentryALTinterwordspacing
M.~Ando, Y.~Chokushi, K.~Tanaka, and K.~Yanagihara, ``Leveraging image-based
  prior in cross-season place recognition,'' in \emph{{IEEE} International
  Conference on Robotics and Automation, {ICRA} 2015, Seattle, WA, USA, 26-30
  May, 2015}, 2015, pp. 5455--5461. [Online]. Available:
  \url{https://doi.org/10.1109/ICRA.2015.7139961}
\BIBentrySTDinterwordspacing

\bibitem{faster-r-cnn}
S.~Ren, K.~He, R.~Girshick, and J.~Sun, ``Faster r-cnn: Towards real-time
  object detection with region proposal networks,'' in \emph{Advances in neural
  information processing systems}, 2015, pp. 91--99.

\bibitem{zeng20163dmatch}
A.~Zeng, S.~Song, M.~Nie{\ss}ner, M.~Fisher, J.~Xiao, and T.~Funkhouser,
  ``3dmatch: Learning local geometric descriptors from rgb-d reconstructions,''
  in \emph{CVPR}, 2017.

\bibitem{icp}
P.~J. Besl and N.~D. McKay, ``Method for registration of 3-d shapes,'' in
  \emph{Sensor Fusion IV: Control Paradigms and Data Structures}, vol.
  1611.\hskip 1em plus 0.5em minus 0.4em\relax International Society for Optics
  and Photonics, 1992, pp. 586--607.

\bibitem{macqueen1967some}
J.~MacQueen \emph{et~al.}, ``Some methods for classification and analysis of
  multivariate observations,'' in \emph{Proceedings of the fifth Berkeley
  symposium on mathematical statistics and probability}, vol.~1, no.~14.\hskip
  1em plus 0.5em minus 0.4em\relax Oakland, CA, USA, 1967, pp. 281--297.

\bibitem{olufs2011robust}
S.~Olufs and M.~Vincze, ``Robust single view room structure segmentation in
  manhattan-like environments from stereo vision,'' in \emph{2011 IEEE
  International Conference on Robotics and Automation}.\hskip 1em plus 0.5em
  minus 0.4em\relax IEEE, 2011, pp. 5315--5322.

\bibitem{sii2014}
H.~{Shogo} and T.~{Kanji}, ``M2t: Local map descriptor,'' in \emph{2014
  IEEE/SICE International Symposium on System Integration}, Dec 2014, pp.
  210--215.

\bibitem{IPIN17}
\BIBentryALTinterwordspacing
E.~Liu, K.~Tanaka, and X.~Fei, ``Grammar-based map parsing for view invariant
  map descriptor,'' in \emph{2017 International Conference on Indoor
  Positioning and Indoor Navigation, {IPIN} 2017, Sapporo, Japan, September
  18-21, 2017}, 2017, pp. 1--8. [Online]. Available:
  \url{https://doi.org/10.1109/IPIN.2017.8115884}
\BIBentrySTDinterwordspacing

\bibitem{DBLP:conf/itsc/TakahashiTF17}
\BIBentryALTinterwordspacing
Y.~Takahashi, K.~Tanaka, and Y.~Fang, ``Cross-season vehicle localization using
  bag of local 3d features,'' in \emph{20th {IEEE} International Conference on
  Intelligent Transportation Systems, {ITSC} 2017, Yokohama, Japan, October
  16-19, 2017}, 2017, pp. 1--6. [Online]. Available:
  \url{https://doi.org/10.1109/ITSC.2017.8317606}
\BIBentrySTDinterwordspacing

\bibitem{spm}
S.~Lazebnik, C.~Schmid, and J.~Ponce, ``Beyond bags of features: Spatial
  pyramid matching for recognizing natural scene categories,'' in \emph{2006
  {IEEE} Computer Society Conference on Computer Vision and Pattern
  Recognition}, 2006, pp. 2169--2178.

\bibitem{rusu2009fast}
R.~B. Rusu, N.~Blodow, and M.~Beetz, ``Fast point feature histograms (fpfh) for
  3d registration,'' in \emph{2009 IEEE International Conference on Robotics
  and Automation}.\hskip 1em plus 0.5em minus 0.4em\relax IEEE, 2009, pp.
  3212--3217.

\end{thebibliography}

\end{document}